%% file: main.tex
\documentclass{article}

\usepackage[preprint,nonatbib]{neurips_2020}

\usepackage[utf8]{inputenc} 
\usepackage[T1]{fontenc}    
\usepackage{hyperref}       
\usepackage{url}            
\usepackage{booktabs}       
\usepackage{amsfonts}       
\usepackage{nicefrac}       
\usepackage{microtype}      
\input{pakete}
\input{befehle}

\input{makros}

\usepackage[linesnumbered,ruled]{algorithm2e}

\usepackage{hyperref}
\usepackage{algorithmicx}
\title{Gradient flow encoding with distance optimization adaptive step size}

\author{Kyriakos Flouris \\ 
Computer Vision Laboratory, ETH Zurich \\
\texttt{kflouris@vision.ee.ethz.ch}
\And
Anna Volokitin \\
Computer Vision Laboratory, ETH Zurich \\
\texttt{voanna@vision.ee.ethz.ch}
\AND
Gustav Bredell \\
Computer Vision Laboratory, ETH Zurich \\
\texttt{gustav.bredell@vision.ee.ethz.ch}
\AND
Ender Konukoglu \\
Computer Vision Laboratory, ETH Zurich \\
\texttt{ender.konukoglu@vision.ee.ethz.ch}
\AND
}

\begin{document}

\maketitle

\begin{abstract}
The autoencoder model uses an encoder to map data samples to a lower dimensional latent space and then a decoder to map the latent space representations back to the data space. Implicitly, it relies on the encoder to approximate the inverse of the decoder network, so that samples can be mapped to and back from the latent space faithfully. This approximation may lead to sub-optimal latent space representations. In this work, we investigate a decoder-only method that uses gradient flow to encode data samples in the latent space. The gradient flow is defined based on a given decoder and aims to find the optimal latent space representation for any given sample through optimisation, eliminating the need of an approximate inversion through an encoder. Implementing gradient flow through ordinary differential equations (ODE), we leverage the adjoint method to train a given decoder. We further show empirically that the costly integrals in the adjoint method may not be entirely necessary. Additionally, we propose a $2^{nd}$ order ODE variant to the method, which approximates Nesterov's accelerated gradient descent, with faster convergence per iteration. Commonly used ODE solvers can be quite sensitive to the integration step-size depending on the stiffness of the ODE. To overcome the sensitivity for gradient flow encoding, we use an adaptive solver that prioritises minimising loss at each integration step. We assess the proposed method in comparison to the autoencoding model. In our experiments, GFE showed a much higher data-efficiency than the autoencoding model, which can be crucial for data scarce applications.
\end{abstract}

\section{Introduction}

Auto-encoders are wildly successful as artificial neural network architectures for unsupervised learning \cite{stackedAE}. The appeal is centred on learning a lower dimensional representation of the input data allowing extremely efficient computation. The idea is simple as it is brilliant, provided there is correlation between the input data the latent dimension can be leveraged to output a model of the input. Nevertheless, the encoding process is semi-arbitrary as there is no direct learning of the encoder, rather than the decoder is optimised to rectify the encoding process. This sub-optimal latent space representations can lead to inefficient learning, \cite{deepsdf}. In part to compensate for this learning process extensive work has been done to regularise the latent space more directly, \cite{Tschannen2018RecentAI}.

Auto-encoders depend on the encoder to approximate the inverse of the decoder network, this approximate inversion requires to be learned with additional parameters and therefore may get in the way of learning with fewer data and adversely affect the latent space structure. Some research in flow models, \cite{DBLP:journals/corr/DinhKB14}, can resolve this issue by using invertible maps however they are constrained to equi-dimensional latent representations. What if we can eliminate the need of an encoder neural network altogether, retain the advantage of lower-dimensional latent space, while obtaining a directly optimised representation?  This may allow us to create a model that can map images to lower dimensional latent space and reconstruct faithfully with fewer images and less iterations.



In this work we propose a novel encoding method namely a gradient flow encoding (GFE). This decoder-only method at each training step primarily determines the optimal latent space representation for each sample via a gradient flow, namely, an ordinary differential equation (ODE) solver. The decoder is updated as usual by minimising the loss between input image and its reconstruction retrieved from the optimal latent space representation for the image. The method, albeit being slower, is considerably superior in data efficiency and ultimately obtains better reconstructions with very few number of training samples when compared to a conventional auto-encoder (AE). A side-advantage of GFE is 'halving' the size of the AE neural network, no need for an encoder.


Using a gradient flow for encoding can be a computationally challenging task. Traditionally, ODE solvers with adaptive step size are propagating the variables in such a manner as to minimise the error of integration. These solvers are crucial for the accuracy of the ODE solution. However, they also impede training of ODE based neural networks if the underlying ODE defined by the network becomes stiff, \cite{chen2018neural, grathwohl2019ffjord}. Adaptive step size becomes too small and integration takes long. When using GFE at each training step, using an ODE solver with adaptive step size can be a debilitating factor. 

For integrating gradient flow in an encoding scheme, we observe that the exact path of the gradient flow may be less important than the final point of convergence. Therefore, adapting step size to minimise error in integration is not necessarily the best approach. Using a fixed step size is also not an option as it may result in poor convergence and stability issues in training the decoder network. 
Here, we develop an adaptive solver for the gradient flow, which adapts each optimisation ('time') step such as to directly minimise the distance, i.e. loss between input and output. Furthermore, this adaptive minimise distance (AMD) 
solver is modified to include a loss convergence assertion to improve performance. The AMD method can possibly make gradient flow a computationally feasible module to be used in neural network architectures. 

Here, an adjoint method, \cite{butcher2008numerical}, as also used in other recent works \cite{chen2018neural}, is implemented to properly optimise the latent space and decoder of the GFE. Consequently, for efficiency considerations we show that a full adjoint method is not necessarily needed and an approximation is utilised. Furthermore, we present the implementation of a Nesterov $2^{nd}$ order ODE solver with accelerated convergence per training data size. Ultimately, the approximate GFE utilising AMD (GFE-amd) is employed for testing and comparison with a traditional AE solver.  

\section{Relevant works}


\textbf{DeepSDF: Learning Continuous Signed Distance Functions for Shape Representation} \cite{deepsdf}: The authors replace the auto-encoder network with an auto-decoder where, a similar to here, latent vector $z$ is introduced. $z$ represents the encoding of a desired shape. They map this latent vector to a 3D shape represented
by a continuous signed Distance Function. I.e. For shape $i$ with function $f_\t$ and coordinate $\mathbf{x}$, $f_\t(z_i,\mathbf{x}) \approx SDF^i(\mathbf{x})$. By conditioning of the network output on a latent vector, they model multiple SDFs with a single
neural network.  




\section{Method}
 An auto-encoder funnels an input $y$ into a lower dimensional latent space representation $z$ using an encoder network $E$ and reconstructs it back using a decoder network $D$. Here, the encoder $E$ and decoder $D$ networks can be thought as approximate inverses of each other. During training, each sample is mapped to the latent space using $E$, mapped back using $D$ and the reconstruction error is minimised with respect to the parameters of both of the networks. 
 
 At any point in this training process, given the decoder an optimal latent space representation for each sample can be defined as the $z^*$ that minimises the reconstruction error. The encoder however, does not map the sample to that optimal $z^*$. So, instead of trying to get the reconstruction of $z^*$ to get closer to the sample, its parameters are updated to get the reconstruction of $E(z)$ to get closer to the sample. This may not be the most efficient use of data samples. Alternatively, one can determine $z^*$ for each sample and update the decoder's parameters according to this optimal latent representation. 
 
 
 Determination of $z^*$ for each sample $y$ can be formulated as an optimisation problem
 \begin{equation*}
     z^* = \arg_z\min l(y, D(z,\th)),
 \end{equation*}
 where $\theta$ represent the parameters of the decoder network and $l(\cdot,\cdot)$ is a distance function between the sample and its reconstruction by the decoder, which can be defined based on the application. One obvious form can be the $L_2$ distance $\lVert D(z,\th) - y \rVert^2_2$.
 The optimisation can be achieved by a gradient decent minimisation. 
 In order to integrate the minimisation in the training of the decoder network, a continuous gradient decent algorithm is implemented via a solution to an ordinary differential equation $d z/d t = -\a(t)\nabla_z l(y, D(z(t),\theta))$,  where time $t$ is the continuous parameter describing the extent of the minimisation and $\a(t)$ is a scaling factor that can vary with time. When the extremum is reached the $z$ comes to a steady state. In practice we compute the optimal $z^*$ by integrating the ODE
 \begin{equation*}
    z^* = z(\t) = \int_0^{\t} -\a(t)\nabla_z l(y, D(z(t), \th)) dt,\ z(0) = z_0,
 \end{equation*}
 where $z_0$ is the initialisation of the optimisation, which is set as 0 vector in our experiments. 
 Consequently to the minimisation of $z \rightarrow z^*\equiv z(t=\t)$ for a given $D$ ('forward model'), the decoder is trained with a total loss function for a given training set $M$ ('backward'),
\begin{equation}
    \mathcal{L}(\theta) = \sum_{m=1}^M l(y_m, D(z_m^*,\theta)).
\end{equation}
At each iteration while searching for the $\arg_{\th}\min \mathcal{ L}(\theta)$, a new $z^*$ is recalculated for each sample. $\arg_{\th}\min \mathcal{ L}(\theta)$ is computed via the adjoint method as explained in Section \ref{sec:adjoint}.

\subsection{The adjoint method for the gradient flow}
\label{sec:adjoint}
 As described above, after finding  $z^* \equiv z(\t) =\arg_z\min l(y, D(z,\th))$ the total loss is minimised with respect to the model parameters. 
 The dependence of $z^*$ to $\th$ creates an additional dependence of $\mathcal{L}(\th)$ to $\th$ via $z^*$.
 For simplicity, let us consider the cost of only one sample $y$, effectively $l(y, D(z^*, \th))$. We will compute the total derivative $d_\th l(y, D(z^*, \th))$ for this sample and the derivative of the total cost for a batch of samples can be computed as the sum of the sample derivatives in the batch.
 The total derivative $d_\th l(y, D(z^*, \th))$ is computed as
\begin{equation*}
    d_\theta l(y, D(z^*, \th)) = \partial_\theta l(y, D(z^*, \th)) + \partial_{z^*}l(y, D(z^*, \th))\partial_\theta z^*.
\end{equation*}
The derivative $\partial_{z^*}l(y, D(z^*, \th))d_\theta z^*$ can be computed using the adjoint method and leads to the following set of equations
\begin{align}
\label{eq:adjoint1}
d z/d t = - \a(t) \nabla_z l(y, D(z(t), \th)), \text{ with } z(0)=0 \\
\label{eq:adjoint2}
d\l/d t  = - \a(t) \l^T  \nabla^2_z l(y, D(z(t), \th)), \text{ with } \l(\tau) = -\nabla_z l(y, D(z(\t), \th))\\ 
\label{eq:adjoint3}
d_\th l(y, D(z^*, \th))= \del_\th l(y, D(z^*, \th))
- \int_0^{\t} \a(t) \lambda^{T} \del_\th  \nabla_z l(z(t),\th) dt, 
\end{align}
where we used $z^* = z(\t)$. 
Equations [\ref{eq:adjoint1}-\ref{eq:adjoint3}] define the so called adjoint method for gradient flow optimisation of the loss. 
Due to the cost of solving all three equations we empirically find that for this work sufficient and efficient optimisation can be accomplished by ignoring the integral (``adjoint function'') part of the method. Theoretically, this is equivalent to ignoring the higher order term of the total differential $d_\th l(y, D(z^*, \th))= \del_\th l(y, D(z^*, \th)) + \del_z l(y, D(z^*, \th)) \del_\th z \approx \del_\th l(y, D(z^*, \th))$. Reducing Equations [\ref{eq:adjoint1}-\ref{eq:adjoint3}] to 
\begin{align}
\label{eq:approx1}
d z/d t = - \a(t) \nabla_z l(y, D(z(t), \th)), \text{ with } z(0)=0 \\
\label{eq:approx2}
d_\th \mathcal{L}= \del_\th l(y, D(z^*, \th)),
\end{align}
i.e. optimise the latent space via solving an ordinary differential equation and minimise the loss ``naively'' with respect to the parameters ignoring the dependence of $z(\t)$ to $\th$. 
\subsection{Nesterov 2-nd order accelerated gradient flow}
\label{sec:nesterov}

The gradient flow described above is based on naive gradient descent. This method may be slow in convergence. The convergence per iteration can be further increased by considering Nesterov's accelerated gradient descent. A second differential equation approximating Nesterov's accelerated gradient method has been developed in \cite{nesterov2016}. This $2^{nd}$ order ODE equation for $z$ reads
\begin{equation}
    \frac{d^2 z}{dt^2}+\frac{3}{t}\frac{dz}{dt}+\nabla_zl(y, D(z, \th))=0
\end{equation}
for $dz/dt|_0=z(0)=0$. To be able to use this in the framework of the gradient flow encoding we split the $2^{nd}$ order ODE into two interacting  $1^{st}$ order equations and solve the simultaneously. Specifically, solving 
\begin{equation}
   \frac{dv}{dt}= \frac{3}{t+ \epsilon}v+\nabla_z l(y, D(z, \th)), \ \ \ \ \ \ \ \frac{dz}{dt}= v,
\end{equation}
where $\epsilon$ ensures stability at small $t$. 

\subsection{Fixed grid ODE solver}
\begin{figure}[ht]
\begin{center}
\includegraphics[width=0.5\columnwidth]{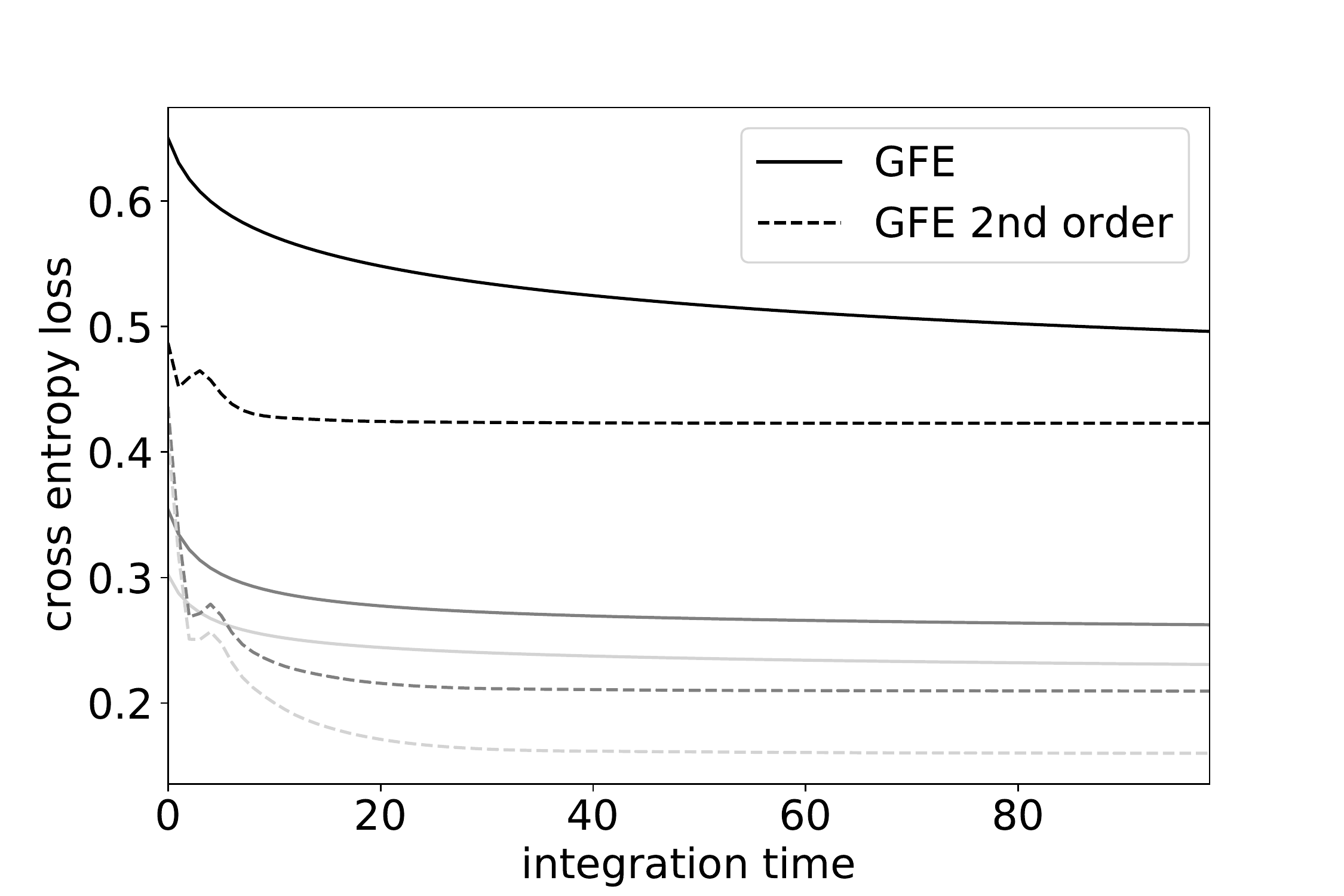}
\caption{\label{fig:conv} Typical convergence curve of $z$ during an ODE integration for GFE and GFE $2^{nd}$ order using the fixed grid ODE solver. Experiments were performed on MNIST. Mean cross-entropy loss plotted against time for three training iterations, darker corresponds to earlier iteration. Curves are obtained using the same batch of training images. The plot shows how convergence curves of $z$ changes with training iterations. Darker curves are for earlier iterations. In later iterations, the convergence curves reach lower loss, meaning updates of the decoder network's parameters lead to more faithful latent space representations, as expected.}
\end{center}
\end{figure}

Being an ODE, the gradient flow can be solved with general ODE solvers. Unfortunately, generic adaptive step size solvers are not useful because the underlying ODE becomes stiff quickly during training, the step size is reduced to extremely small values and the time it takes to solve gradient flow ODEs at each iteration takes exorbitant amount of time. Fixed time-step or time grid solvers can be used, despite the stiffness. However, we empirically observed that these schemes can lead to instabilities in the training, see Figure~\ref{fig:all_methods}. To demonstrate this, we experimented with a $4^{th}$ order Runge-Kutta method with fixed step size. The $\d t$ slices are predefined in logarithmic series such as $\d t$ is smaller closer to $t=0$, where integrands, $-\nabla_z l(y,D(z,\th))$, are more rapidly changing. Similarly, $\a$ is empirically set to $e^{(-2t/\t)}$ to facilitate faster convergence of $z$, see Figure \ref{fig:conv}. For the GFE full adjoint method the integrands for each time slice are saved during the forward pass so they can be use for the calculation of the adjoin variable $\l$ in the backward pass. We used the same strategy for both basic gradient flow and the $2^{nd}$ order model.

\subsection{Adaptive minimise distance (AMD) solver}
\label{sec:amd}
The step size in any optimisation algorithm is of fundamental importance for reaching a local extremum in any optimisation problem. 
The same applies to the search for optimal latent space representation for each sample by solving gradient flow equation through its time discretisation.
As discussed, fixed time step or time grid solvers empirically lead to instabilities during training of the decoder. 
Adaptive step size ODE solvers can theoretically solve the instabilities but due to the possible stiffness of the gradient flow equation, their use is impeded. 
Note that, the adaptive step size solvers focus on an accurate integration, which is not necessarily an advantage for integrating gradient flow in the training of a decoder. A solver that focuses on reaching a local extremum while searching for the optimal $z$ would be more advantageous. The path it takes to reach there may be less crucial as long as it reaches the same local extremum. This is in contrast to generic ODE solvers where the path can substantially change the end point, thus an focusing on accurate integration is more important. To this end, we develop an adaptive step size method, which guarantees a reduction of the loss at each step. 

The method follows a similar structure to an explicit ODE solver, such as the feed-forward Euler method, but without a fixed grid. The problem lies in solving Equation \ref{eq:approx1} while taking time-steps of appropriate size that reduce $l(y,D(z(t),\th))$ at each $t$. 
This approach is in essence a gradient descent method that uses step-size selection mechanisms~\cite{bertsekas_2016}. Lastly, viewing the time-step in solving $d{z}/dt = -\a(t) \nabla_z l(y, D(z(t),\theta))$ as a tool to minimise the loss makes $\a(t)$ obsolete. Its role is now overtaken by the time-step $\d t$ and it can be set to 1 for all t.

 In the AMD method, at each time $t$ the time step is chosen based on finding the smallest $m=0,1,\dots,$ that satisfies
 \begin{equation}
     l\left(y,D(z(t_n) - \beta^m s_n \nabla_z l(y, D(z(t_n),\theta)), \th)\right) < l\left(y,D(z(t_n), \th)\right)
 \end{equation}
 with $\beta\in(0,1)$ (set as 0.75 in our experiments) and $s_n$ is a scaling factor. At each time point $t_n$ the time step is chosen as $\d t_n = \beta^m s_n$. The scaling factor is updated at each iteration as
 \begin{equation*}
     \hat{s}_n = \max(\kappa s_{n-1}, s_0), \ s_n = \min(\hat{s}_n, s_{max}), \ s_{max}=10,\ s_0 = 1,\ \kappa = 1.1 
 \end{equation*}
 Based on this $t_{n+1} = t_n + \d t_n$ and
 \begin{equation}
     z(t_{n+1}) = z(t_n) - \d t_n \nabla_z l(y, D(z(t_n),\theta)
 \end{equation}


At the end, if time step chosen goes beyond $\tau$, a smaller time step is used to reach $\tau$ exactly. The solution of the integral \ref{eq:approx1} is then $z(\tau)$. Furthermore, the AMD solver is using the gradient of the convergence curve (see Figure \ref{fig:conv}) to assert if the loss function is sufficiently optimised to assign a new final $\tau'$ and stop in order to avoid unnecessary integration.

\begin{algorithm}[tb]
   \caption{training GFE-amd for one sample per batch}
   \label{alg:example}
\begin{algorithmic}
   \State {\bfseries Input: } z(0)
   \State {\bfseries Output:} {$\mathbf{y}_b$, $\mathbf{\theta}_D$, $\mathbf{z}^*_b$} 
   \For{$b \in \{1,2,...,B\}$}{
   \While {$\frac{d}{db} {L}_{CE}(b) < 0.01$}{
   \While {$l(z({t}^n_{b} + \delta {t}^n_{b}) < l(z({t}^n_{b}$))}{
   \State $z({t}^{n+1}_{b}) \leftarrow z({t}^{n}_{b}) + [-\nabla_z l(z({t}^{n}_{b}))]\delta {t}^n_{b}$
   \State {$\delta {t}^{n+1}_{b} \leftarrow \beta^r\delta {t}^n_{b}$}
   \State {$r \leftarrow r+1$}
    }
    \State {$n \leftarrow n+1$}
   }
   \State {$z^*_b \leftarrow z({t}^n_{b})$}
   \State {$L(\theta_D) \leftarrow l(y, D(z^*, \theta_D)$}
   \State {$\theta_D \leftarrow$ backpropagate $L(\theta_D)$}
    }
\end{algorithmic}
\end{algorithm}

\section{Experimental setup}

\begin{figure}[ht]
\begin{center}
\includegraphics[width=0.4\columnwidth]{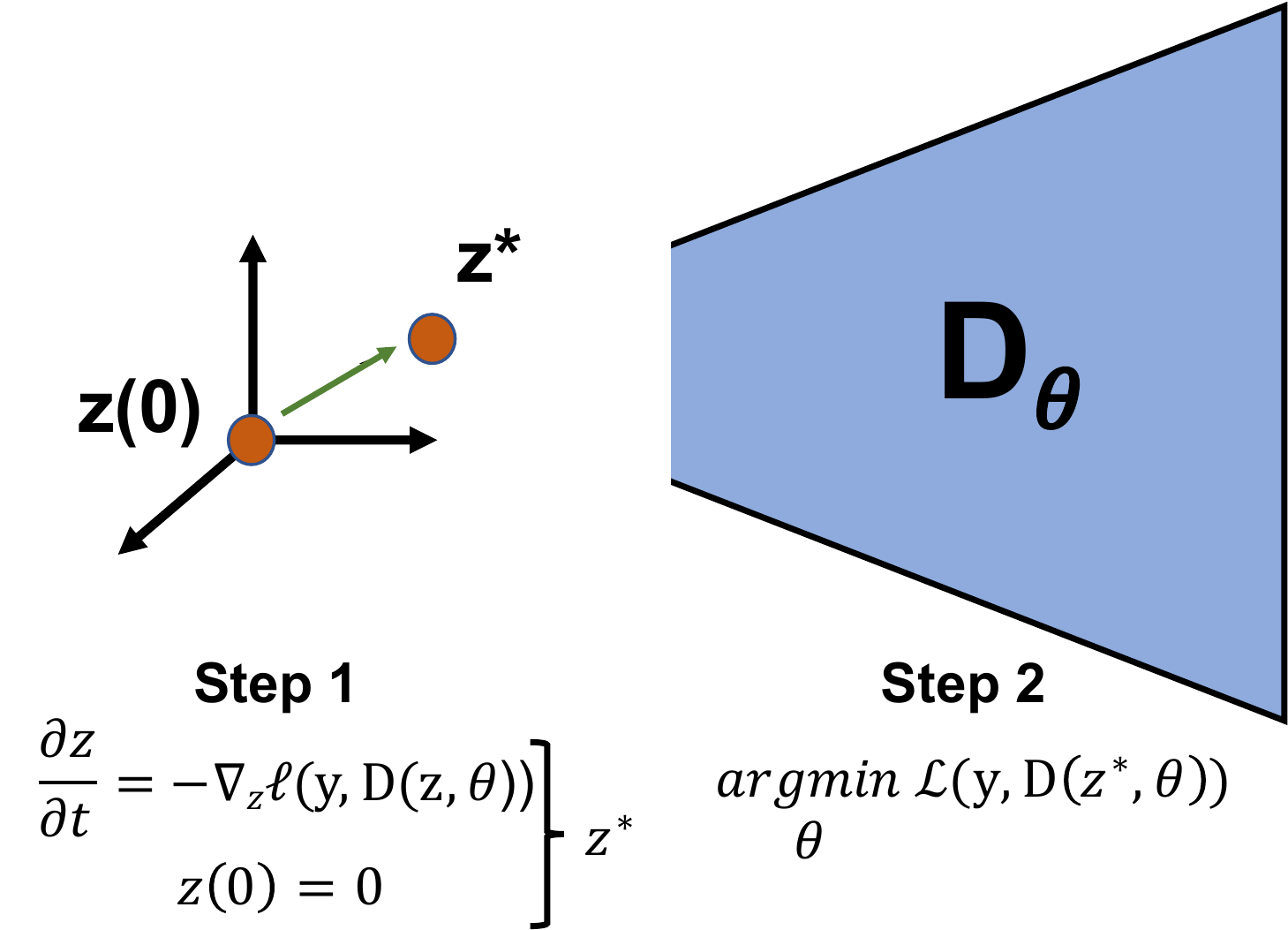}
\caption{\label{fig:schematic} Schematic diagram of the GFE implementation. The two optimisation correspond to (Step 1) the latent space optimisation i.e. 'encoding' and  (Step 2) the parameter update of the neural network.}
\end{center}
\end{figure}

For training with MNIST and FashionMNIST datasets we implement a sequential linear neural network. The decoder network architecture corresponds to four gradually increasing linear layers with ELU non-linearities in-between. The exact reverse is used for the encoder of the AE. A schematic diagram of the GFE method is shown in Figure \ref{fig:schematic}.

The network training is carried out with a momentum gradient decent optimiser (RMSprop), learning rate $0.0005$, $\e=1 \times 10^{-6}$ and $\a=0.9$. The GFE and AE are considered trained after one epoch and twelve epochs respectively. 

\section{Results and Discussion}

\begin{figure}[ht]
\begin{center}
\includegraphics[width=1\columnwidth]{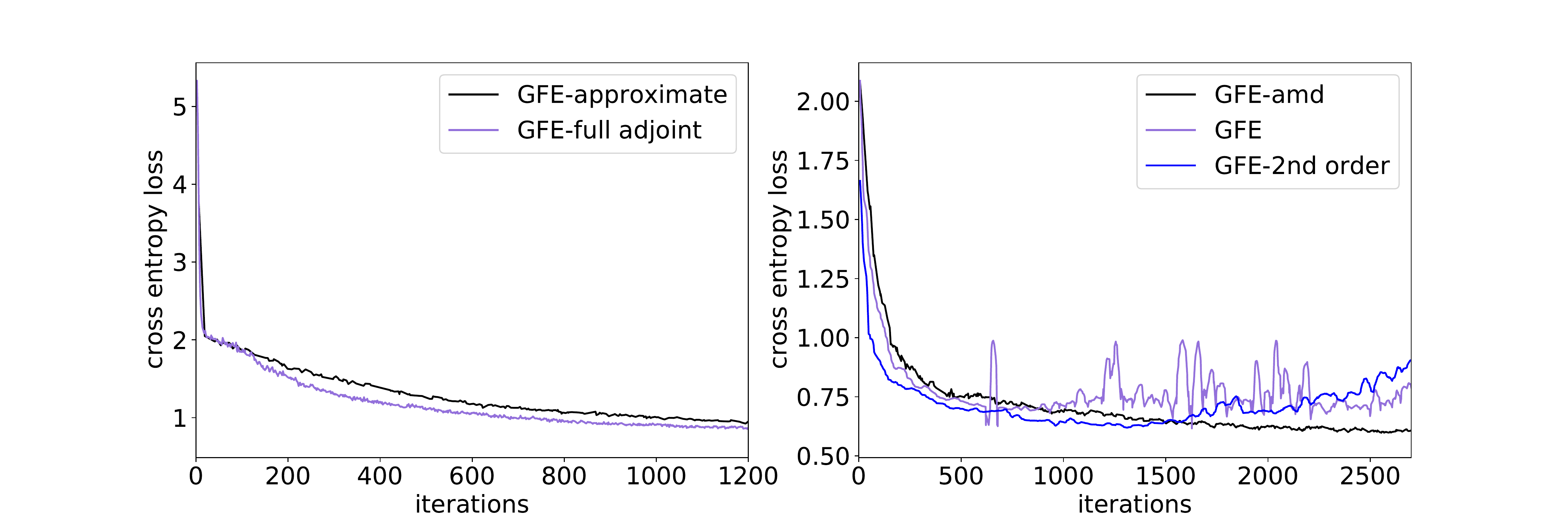}
\caption{\label{fig:all_methods} \textbf{Left} Validation mean cross-entropy loss plotted against MNIST training iterations for the approximate and full adjoint GFE methods. The full adjoint has a slight advantage over the approximate. \textbf{Right} Validation mean cross-entropy loss plotted against MNIST training iterations for the GFE, $2^{nd}$ order GFE and GFE-amd methods. The GFE-amd is both more stable and approaches a better convergence relative to the other methods}
\end{center}
\end{figure}

Initially a relative comparison between the full adjoint and the approximate fixed grid GFE methods is carried out to assess the relevance of the higher order term. Specifically, we carry out MNIST experiments for a fixed network random seed, where we  trained the Decoder using the different GFE methods and computed cross entropy loss over the validation set. The proper adjoint solution requires Equations \ref{eq:adjoint1} and \ref{eq:adjoint2} to be solved for each slice of the integral in Equation \ref{eq:adjoint3}. Given $N$ time-slices (for sufficient accuracy $N\approx 100$), this requires $\mathcal{O}(5N)$ calls to the model $D$ for each training image.  The approximate method as in Equations \ref{eq:approx1} and \ref{eq:approx2} requires only $\mathcal{O}(N)$ passes. From Figure \ref{fig:all_methods}~(left) it is evident that the 5-fold increase in computational time is not cost-effective as the relative reduction in loss convergence per iteration is not significant. 

Furthermore, to increase convergence with respect to training data the accelerated gradient flow $2^{nd}$ order GFE is implemented in Section \ref{sec:nesterov}.
From Figure \ref{fig:all_methods}~(right),  the accelerated gradient method increases initially the convergence per iteration relative to GFE, nevertheless it is slightly more computationally expensive due to solving a coupled system. Additionally, from the same Figure certain stability issues are observed for both GFE and second order GFE methods later on despite the initial efficient learning. In order to guarantee  stability the GFE-amd method is implemented as explained in Section \ref{sec:amd}. The black curve in Figure \ref{fig:all_methods}~(right) shows a clear improvement of the GFE-amd over the later methods. Importantly, this result is robust to O of the experiment.

\begin{figure}[ht]
\begin{center}
\includegraphics[width=\columnwidth]{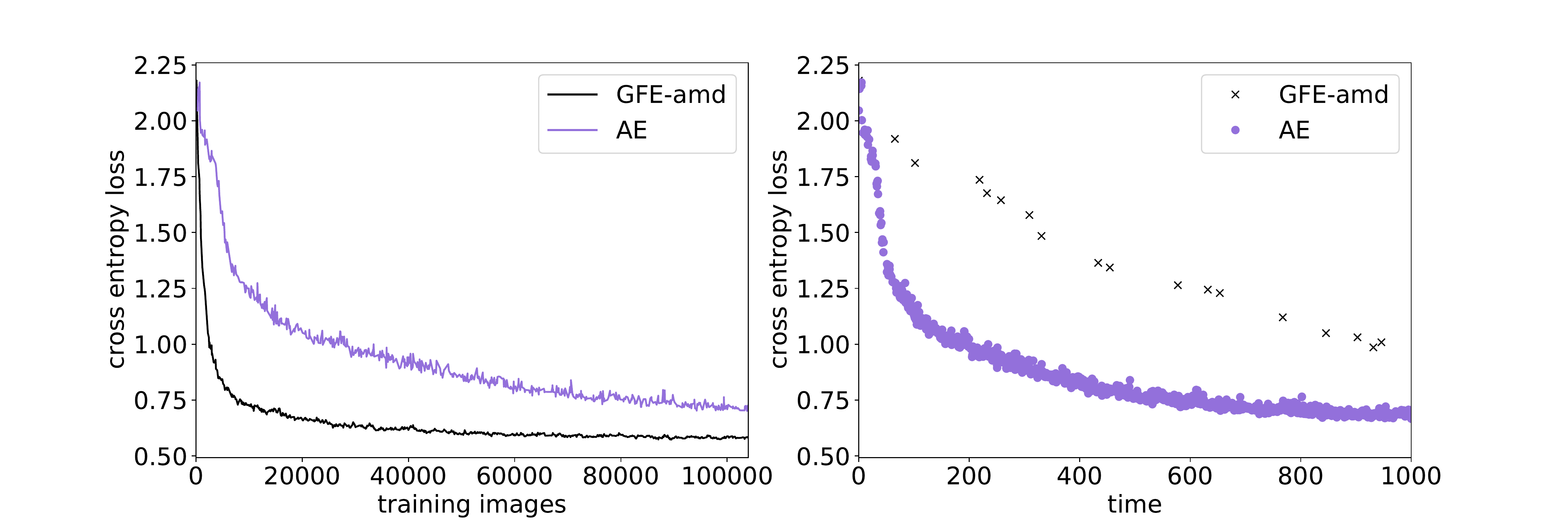}
\end{center}
\caption{\label{fig:GFE_AE_step_relative} \textbf{Left} Validation mean cross-entropy loss plotted against number of MNIST training images for the GFE-amd and AE methods. The former shows a significant convergence with a tiny amound of training images. \textbf{Right} Validation mean cross-entropy  loss  plotted  against  time  for  the  GFE-amd and AE methods. The latter is significantly faster to the former as much more iterations are carried out in the same time-span}
\end{figure}

A direct comparison of the GFE-amd to a conventional AE for an MNIST training can be seen in Figure \ref{fig:GFE_AE_step_relative}. The x-axis shows the number of training images (instead of iterations) the algorithm sees until that point in the training. The training is based on mini-batch training using the data with replacement, going over the training data multiple times. The GFE-amd is substantially superior in learning per training image, reaching near convergence with at 800000 images, see Figure \ref{fig:GFE_AE_step_relative}~(left). This is a consequence of the efficiently optimised latent space. Nevertheless, this comes at a  higher computational cost for each iteration due to the ODE solver as seen from Figure \ref{fig:GFE_AE_step_relative}~(right). Importantly, the optimisation of the network parameters is performed using Adam optimiser for both AE and GFE models. So the difference we see can be attributed to better gradients GFE model generates to update the decoder network at each training iteration.

This increase in computation is not necessarily a disadvantage considering the efficient learning of the GFE-amd method. In Table \ref{tb:trainingimages}, the average cross entropy loss for a complete test-set is recorded for both methods for some small number of training images. The GFE-amd is able to learn quite well even after seeing a tiny fraction of the total training data. Furthermore, the GFE-amd method noticeably improves an AE trained decoder when it is used to test, the result of an optimised latent space even without a network parameters update.


\begin{table}[ht]
\centering
 \begin{tabular}{||c c c c||} 
 \hline
Number of Training images & AE & GFE-amd  & train:AE test:GFE-amd
\\ [0.5ex] 
 \hline\hline
 480 (0.24$\%$)  & 0.2660 & 0.2098  & 0.2634\\ 
 \hline
 960 (0.49$\%$)  & 0.2618 & 0.1987 & 0.2525 \\ 
 \hline
 1920 (0.98$\%$) & 0.2488 & 0.1558 & 0.2323 \\
 \hline
 3840 (1.95$\%$) & 0.2195 & 0.1336 & 0.2038 \\
 \hline
 5760 (2.9$\%$) & 0.1954 & 0.1136 & 0.1829 \\ 
 \hline
\end{tabular}
\caption{\label{tb:trainingimages} Test-set average cross entropy loss for different number of training data, $\%$ reflects the percentage relative to total training data needed for convergence of the method. Carried out for the AE GFE-amd methods. The forth column represents testing the decoder of an AE trained network with GFE-amd. The GFE-amd is far superior here in learning with a limited image sample.}
\end{table}

To verify the overall quality of the method both the AE and GFE-amd are tested when converged as shown in Table \ref{tb:loss}. 
The GFE-amd performs very similar to AE both for MNIST, SegmentedMNIST and FMNIST. 
It is worth noting that the GFE-amd trainings are on average converged at $1/12^{th}$ of the number of iterations relative to the AE. 
For the segmented MNIST the networks are fully trained while seeing only the first half (0-4) of the MNIST labels and they are tested with the second half (5-9) of the labels. The GFE-amd shows a clear advantage over the AE emphasizing the versatility of a GFE-amd trained neural network.

\begin{table}[ht]
\centering
 \begin{tabular}{||c c c||} 
 \hline
Dataset & AE (Test-set) & GFE-amd (Test-set)
\\ [0.5ex] 
 \hline\hline
 MNIST  & 0.0843 &  0.0830\\ 
 \hline
 SegmentedMNIST & 0.1205 & 0.1135  \\
 \hline
 FMNIST & 0.2752 &  0.2764 \\
 \hline
\end{tabular}
\caption{\label{tb:loss} Test-set average cross entropy loss for trained networks with the AE and GFE-amd methods. Segmented MNIST represents training for half the labels and testing with the other half.}
\end{table}

\begin{figure}[ht]
\begin{center}
\includegraphics[width=\columnwidth]{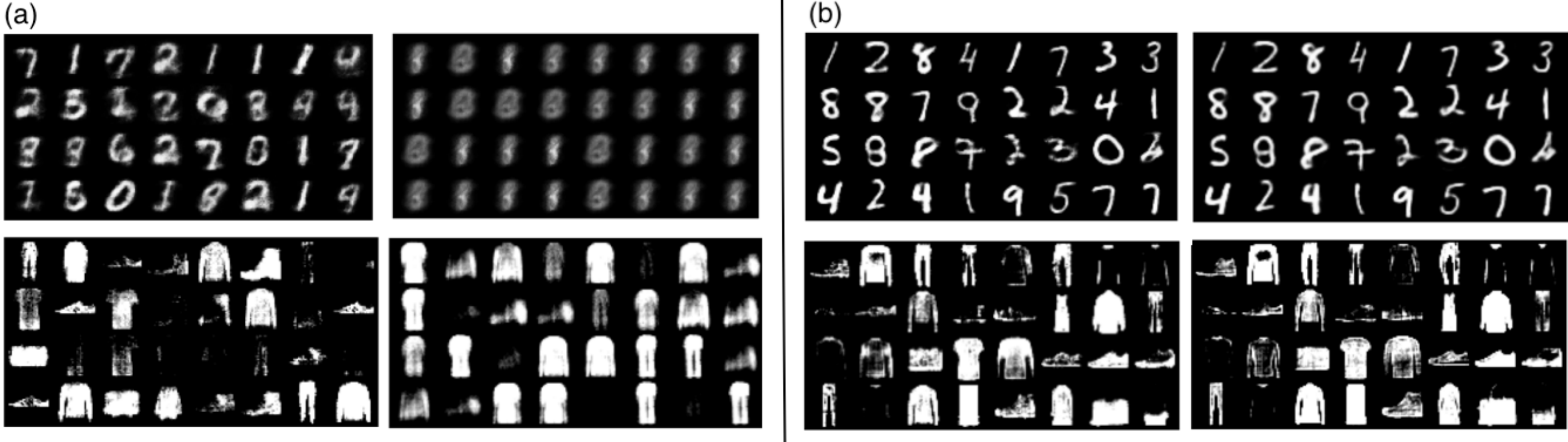}
\end{center}
\caption{\label{fig:images} \textbf{(a)} Test-set reconstructions for trained GFE-amd (left) and AE (right)  that only see  $1\%$ of MNIST (top) and FashionMNIST (bottom) training images. \textbf{(b)} Test-set reconstructions for fully trained GFE-amd (left) and AE (right)  with  MNIST (top) and FashionMNIST (bottom) training images. Note: The labels are identical in the respective reconstructions.}
\end{figure}

Sample test-set reconstructions with a fixed network random seed for GFE-amd and AE methods are shown in Figure \ref{fig:images}. From Figure \ref{fig:images}~(a) it is evident that the GFE-amd is superior in producing accurate reconstructions with the very limited amount of data. Figure \ref{fig:images}~(b) indicates that both GFE-amd and AE generate similar reconstructions when properly trained. 

\begin{figure}[ht]
\begin{center}
\includegraphics[width=0.8\columnwidth]{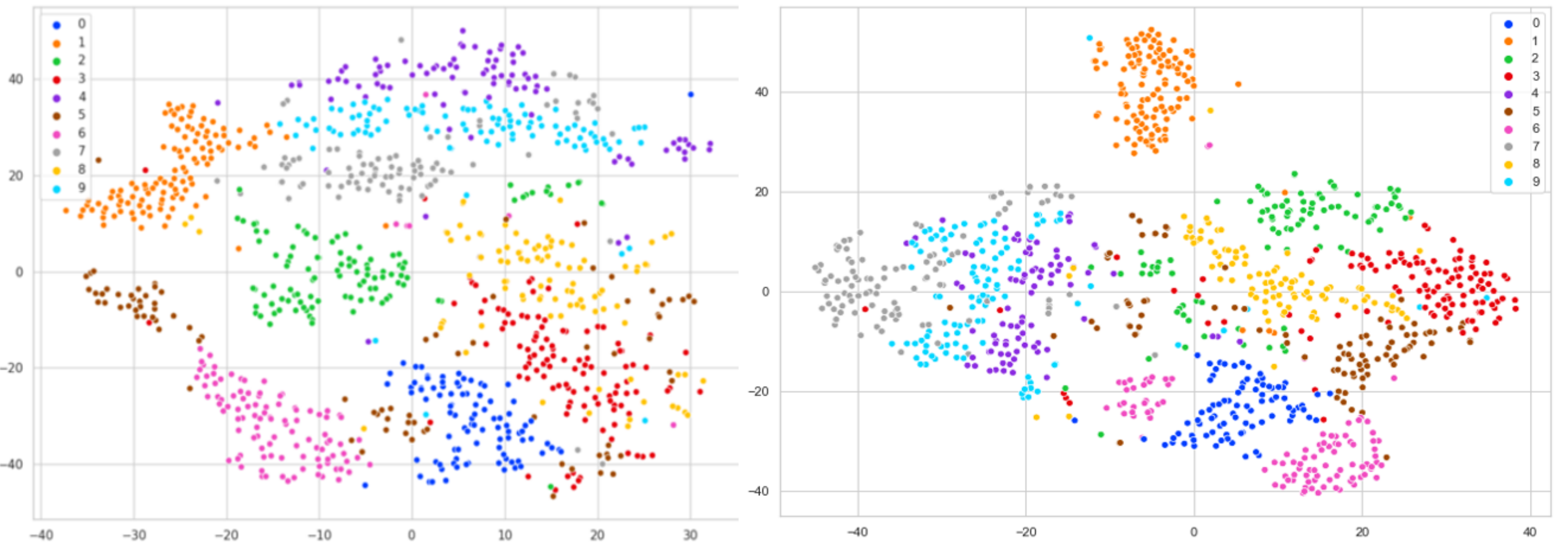}
\end{center}

\caption{\label{fig:tsne} t-SNE map of the latent space plotted for MNIST trained \textbf{Left} GFE and \textbf{Right} AE methods.}

\begin{center}
\includegraphics[width=0.8\columnwidth]{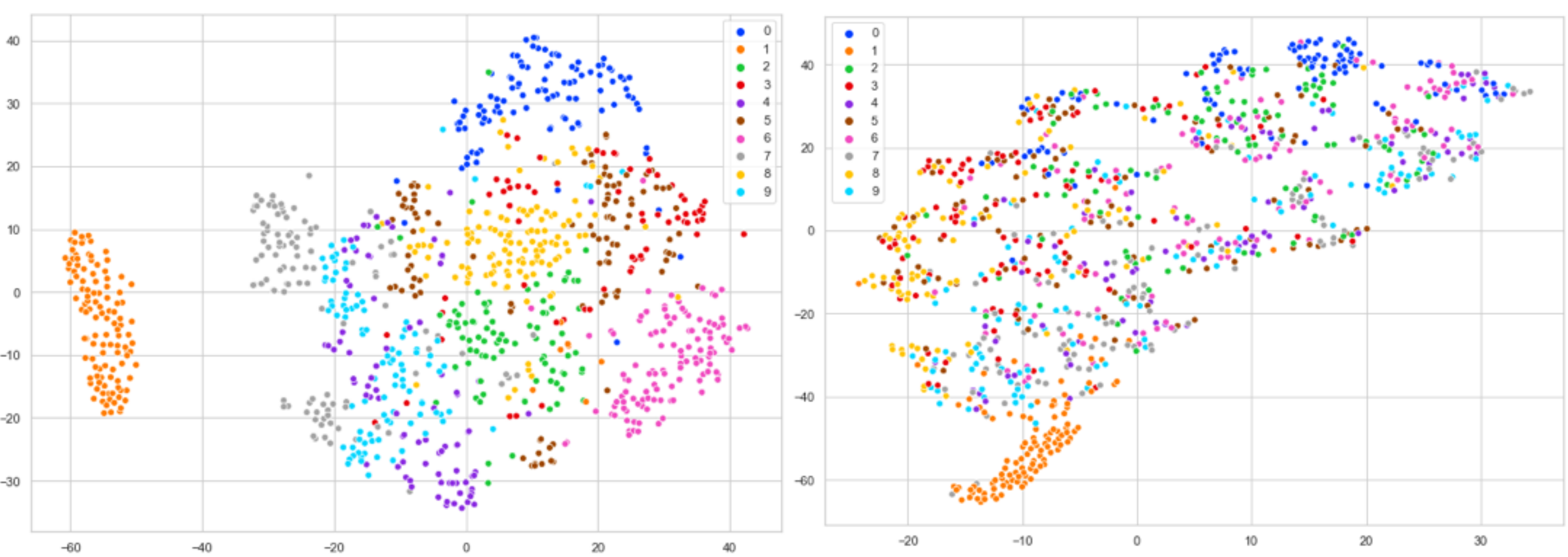}
\end{center}

\caption{\label{fig:tsne_training} t-SNE map of the latent space plotted for MNIST trained with $1\%$ training data \textbf{Left} GFE and \textbf{Right} AE methods. The GFE latent space is well optimised even with a fraction of the data. }
\end{figure}

Finally, to further compare latent space representation, we visualize the samples in the latent space using the t-distributed stochastic neighbour embedding (t-SNE) map, \cite{tsne2008}. This is calculated for the GFE optimised $z \rightarrow z*$. This is shown for AE and GFE,  MNIST trained neural networks in Figure \ref{fig:tsne}. The latent space representations are similar when both models see the entire MNIST dataset multiple times. Similar t-SNE plots for models that only see 1\% of the data during training are given in Figure \ref{fig:tsne_training}. Latent space structure of GFE is very similar for both cases, while AE's latent space structure is very different, not clustering the different numbers. This result is inline with Table~\ref{tb:trainingimages}. GFE uses training images more efficiently at each iteration thanks to latent space optimisation to invert the decoder rather than using an approximation through an encoder network. 


\section{Conclusions}
To this end, a gradient flow encoding, decoder-only method was investigated. The decoder depended gradient flow searches for the optimal latent space representation, which eliminates the need of an approximate inversion. The full adjoint solution and its approximation or leveraged for training and compared. Furthermore, we present  a $2^{nd}$ order ODE variant to the method, which approximates Nesterov's accelerated gradient descent, with faster convergence per iteration. Additionally,  an adaptive solver that prioritises minimising loss at each integration step is described and utilised for comparative tests to the autoencoding model. The gradient flow encoding shows a much higher data-efficiency than the autoencoding model.

\bibliographystyle{abbrv}
\bibliography{main}

\end{document}

%% file: pakete.tex
\usepackage[T1]{fontenc}

\usepackage[ngerman,british]{babel}
	\addto\captionsngerman{}

\usepackage{amsmath,amssymb,amsthm,dsfont,mathtools,mathrsfs,bbm}

\usepackage{mdwlist}
\usepackage{paralist}
\usepackage{array}
\usepackage{booktabs}						
\usepackage{dcolumn} 						

\usepackage{graphicx}
\usepackage{wrapfig}

\usepackage{url}

\usepackage{color}
	\definecolor{myblue}{rgb}{0,0.3,0.8}
	\definecolor{mygreen}{rgb}{0,0.5,0}
	\definecolor{myblue}{rgb}{0,0.3,0.8}

\usepackage[version=3]{mhchem}

\usepackage{savesym}



%% file: befehle.tex
\savesymbol{a}
\savesymbol{b}
\savesymbol{d}
\savesymbol{e}
\savesymbol{k}
\savesymbol{l}
\savesymbol{H}
\savesymbol{L}
\savesymbol{p}
\savesymbol{r}
\savesymbol{s}
\savesymbol{S}
\savesymbol{F}
\savesymbol{t}
\savesymbol{w}
\savesymbol{th}
\savesymbol{div}
\savesymbol{tr}
\savesymbol{Re}
\savesymbol{Im}
\savesymbol{AA}
\savesymbol{SS}
\savesymbol{Si}
\savesymbol{gg}
\savesymbol{ss}

\newcommand{\a}{\alpha}

\newcommand{\e}{\varepsilon}
\newcommand{\d}{\delta}
\newcommand{\del}{\partial}

\newcommand{\l}{\lambda}

\newcommand{\t}{\tau}
\newcommand{\th}{\theta}



%% file: makros.tex
\newlength{\tmplen}

